\documentclass[acmsmall]{acmart}
\AtBeginDocument{%
  }

\setcopyright{none}

\usepackage{graphicx}
\usepackage{amsmath}
\usepackage{amssymb}
\usepackage{hyperref}
\usepackage{tikz}
\usepackage{booktabs}
\usepackage{caption}
\usepackage{array}
\usepackage{multirow}
\usepackage{amssymb} 
\usepackage{pifont}  
\newcommand{\xmark}{\ding{55}} 
\usetikzlibrary{shapes.geometric, arrows, positioning}
\tikzstyle{process} = [rectangle, minimum width=2.5cm, minimum height=1.5cm, text centered, draw=black, fill=blue!20]
\tikzstyle{data} = [rectangle, minimum width=2cm, minimum height=1.2cm, text centered, draw=black, fill=green!20]
\tikzstyle{arrow} = [thick,->,>=stealth]

\begin{document}

\title{Leveraging Foundation Models for Multimodal Graph-Based Action Recognition}

\author{Fatemeh Ziaeetabar}
\authornote{Corresponding author}
\email{fziaeetabar@ut.ac.ir}
\orcid{0000-0003-1159-3588}

\affiliation{%
  \institution{School of Mathematics, Statistics and Computer Science, College of Science, University of Tehran}
  \city{Tehran}
  \country{Iran}
}

\author{Florentin W\"org\"otter }
\email{worgott@gwdg.de}
\orcid{0000-0001-8206-9738}

\affiliation{%
  \institution{Bernstein Center for Computational Neuroscience, Department for Computational Neuroscience, III Physikalisches Institut-Biophysik, Georg-Agust-Universit\"at G\"ottingen, G\"ottingen}
  \city{G\"ottingen}
  \country{Germany}
}


\begin{abstract}
Foundation models have ushered in a new era for multimodal video understanding by enabling the extraction of rich spatiotemporal and semantic representations. In this work, we introduce a novel graph-based framework that integrates a vision-language foundation, leveraging VideoMAE for dynamic visual encoding and BERT for contextual textual embedding, to address the challenge of recognizing fine-grained bimanual manipulation actions. Departing from conventional static graph architectures, our approach constructs an adaptive multimodal graph where nodes represent frames, objects, and textual annotations, and edges encode spatial, temporal, and semantic relationships. These graph structures evolve dynamically based on learned interactions, allowing for flexible and context-aware reasoning. A task-specific attention mechanism within a Graph Attention Network further enhances this reasoning by modulating edge importance based on action semantics. Through extensive evaluations on diverse benchmark datasets, we demonstrate that our method consistently outperforms state-of-the-art baselines, underscoring the strength of combining foundation models with dynamic graph-based reasoning for robust and generalizable action recognition.
\end{abstract}

\maketitle

\section{Introduction}

Manipulation action recognition is a cornerstone of modern computer vision and robotics, underpinning advancements in human-robot interaction~\cite{bharadhwaj2024towards, su2023recent}, assistive systems~\cite{losey2022learning, giammarino2024super}, and industrial automation~\cite{mukherjee2022survey}. These tasks require precise modeling of complex hand-object interactions that unfold across spatial, temporal, and semantic dimensions ~\cite{ziaeetabar2017semantic, ziaeetabar2018recognition, ziaeetabar2018prediction}. The challenge is further compounded in \textit{bimanual} actions, where successful recognition depends on capturing coordinated inter-hand dynamics and object manipulation—capabilities often underrepresented in traditional frame-based video models ~\cite{Ziaeetabar2024, ziaeetabar2024multi}.

In our prior work~\cite{Ziaeetabar2024}, we proposed a hierarchical graph-based architecture (BiGNN) for the recognition and semantic description of bimanual manipulation actions. This framework employed a rule-based scene graph construction pipeline and a three-tiered Graph Attention Network (GAT) to reason over object-, single-hand-, and bimanual-level interactions. While BiGNN demonstrated promising results in multi-level action understanding, its reliance on handcrafted spatial heuristics for graph topology generation imposed significant constraints on adaptability and scalability. Additionally, visual features were extracted using conventional pipelines (e.g., YOLO~\cite{redmon2016you}, OpenPose~\cite{cao2019openpose}), limiting the model's ability to encode fine-grained motion cues and generalize across heterogeneous manipulation contexts.

To overcome these limitations, the present work introduces a novel graph-based framework that leverages large-scale foundation models (FMs) to construct and refine multimodal interaction graphs in a fully learnable and context-sensitive manner. Specifically, we integrate spatiotemporal representations from VideoMAE~\cite{tong2022videomae} and semantic embeddings from BERT~\cite{devlin2019bert} within a dynamically constructed graph architecture. The model dynamically forms edges and applies task-specific attention mechanisms to capture nuanced hand-object interactions and temporal dependencies, enabling robust and context-aware recognition of fine-grained manipulation tasks.

Our key contributions are summarized as follows:

\begin{itemize}
    \item \textbf{Multimodal Representation Learning:} We leverage foundation models to extract rich spatiotemporal (VideoMAE) and semantic (BERT) embeddings, enabling more robust encoding of hand-object dynamics across diverse manipulation scenarios.
    
    \item \textbf{Data-Driven Graph Construction:} We introduce a fully learnable graph construction strategy in which nodes and edges are instantiated based on cross-modal attention signals, replacing heuristic or fixed graph topologies.

    \item \textbf{Task-Aware Relational Reasoning:} Our model integrates task-specific attention mechanisms within GATs to dynamically modulate the importance of spatial, temporal, and semantic relationships central to bimanual coordination.

    \item \textbf{Dynamic Topology Adaptation:} The proposed framework supports graph evolution during inference, enabling fine-grained reasoning and improved generalization to unseen manipulation contexts and interaction patterns.
\end{itemize}

Unlike prior ``FM + GNN'' approaches that rely on static or partially adaptive graph structures ~\cite{liu2025graph, zhu2024llm, franks2025towards}, our framework introduces a fully learnable graph topology, where both connectivity and edge importance are derived from multimodal embeddings. Instead of treating foundation models as fixed feature extractors, we incorporate their outputs into a joint reasoning pipeline that adapts graph structure and relational semantics based on cross-modal cues. This design enables precise modeling of fine-grained bimanual interactions and demonstrates improved generalization across multiple video benchmarks.

The remainder of this paper is structured as follows. Section~\ref{sec:related_work} reviews prior research on multimodal action recognition and graph-based reasoning. Section~\ref{sec:method} details our proposed framework. Section~\ref{sec:results} presents experimental results and comparative evaluations. Finally, Section~\ref{sec:conclusion} discusses key findings and future research directions.

\section{Related Work}
\label{sec:related_work} 

Human action recognition, especially for manipulation and bimanual tasks, has advanced through developments in multimodal learning, foundation models, and graph-based reasoning. Traditional approaches often struggled with the complexity of hand-object interactions~\cite{vrigkas2015review, choi2017robust, verma2024human, worgotter2020humans, ziaeetabar2020using}, while recent work leverages self-supervised learning~\cite{xie2024dynamic} and adaptive graph-based models~\cite{kochakarn2023explainable} to improve spatiotemporal and semantic understanding.

We review three relevant areas: Section~\ref{sec:multimodal_foundation} covers multimodal learning and foundation models; Section~\ref{sec:graph_reasoning} discusses graph-based reasoning; and Section~\ref{sec:integration} examines efforts to integrate foundation models with graph structures—laying the groundwork for our dynamic, context-aware multimodal graph framework.

\subsection{Multimodal Learning and Foundation Models}
\label{sec:multimodal_foundation}

Multimodal learning has advanced human action recognition by integrating visual, textual, and sensory inputs to capture complex hand-object interactions and inter-hand coordination~\cite{nagrani2021attention, cadene2019murel}. Early works combined RGB data with depth or motion cues~\cite{molchanov2016online, karpathy2014large}, while recent methods incorporate semantic modalities—e.g., textual annotations—for deeper contextual understanding and disambiguation of fine-grained actions~\cite{nagrani2021attention}.

Foundation models (FMs) further expand this capability by offering generalizable pretrained architectures across modalities~\cite{liang2024low, qian2024advancing}. CLIP~\cite{radford2021clip} aligns vision and language features, aiding semantic action reasoning, while VideoMAE learns spatiotemporal representations through masked video modeling. Both have shown strong performance in manipulation-focused tasks.

Recent works apply these models to bimanual scenarios. Gao et al.~\cite{gao2024bikvil} introduced Bi-KVIL, a keypoint-based imitation framework leveraging multimodal inputs, and Liu et al.~\cite{liu2024rdt} proposed the Robotics Diffusion Transformer (RDT), a diffusion-based FM tailored for bimanual manipulation. However, aligning heterogeneous modalities and adapting FMs to domain-specific tasks remain open challenges~\cite{li2023efficient}.

\subsection{Graph-Based Reasoning}
\label{sec:graph_reasoning}

Graph-based reasoning has become a key strategy for modeling the structured relationships underlying manipulation and bimanual action recognition. While multimodal features enrich representation, graphs enable explicit encoding of spatial, temporal, and semantic dependencies among entities such as hands, objects, and contextual cues.

Early models like Spatial-Temporal Graph Convolutional Networks (ST-GCN)~\cite{yan2018spatial} modeled skeletal joints for action recognition, followed by Two-Stream Adaptive GCNs (2s-AGCN)~\cite{shi2019two}, which introduced dynamic graph structures that adapt spatial-temporal links to improve generalization. More recent efforts incorporate multimodal signals: Wang et al.~\cite{wang2023egocentric} proposed egocentric spatial-temporal graphs for first-person hand-object modeling, while Lyu et al.~\cite{lyu2023bimanual} extended this to bimanual coordination.

Dynamic and context-aware graph construction has also gained traction. Xie et al.~\cite{xie2024dynamic} introduced adaptive topologies based on interaction cues, and Wu et al.~\cite{wu2023semantic} enriched graph semantics by embedding object and action-level information. These advances underscore the value of learnable, flexible graphs for capturing the complexity of manipulation behaviors.

\subsection{Integration of Multimodal Learning and Graph-Based Reasoning}
\label{sec:integration}

Several recent works have explored integrating multimodal inputs with graph-based reasoning to improve action recognition. Fusion-GCN~\cite{duhme2021fusiongcn} combined RGB, inertial, and skeletal data within a graph framework to enhance multimodal fusion. Li et al.~\cite{li2022representing} proposed MUSLE, modeling videos as discriminative subgraphs to capture fine-grained spatiotemporal patterns. Ramachandram et al.~\cite{ramachandram2017structure} applied graph-induced kernels to optimize structural integration in deep multimodal fusion networks.

Despite these advances, most prior efforts rely on static graph topologies and do not leverage foundation models (FMs) for high-capacity representation learning. Our work addresses this limitation by combining large-scale pretrained FMs with dynamic graph construction and reasoning. Through task-specific attention and adaptive topology refinement, our framework models spatiotemporal and semantic relationships in a context-aware manner, enabling robust recognition of fine-grained bimanual manipulation actions.

\section{Overview of the Proposed Framework}

We model each video sequence as a structured multimodal graph, where nodes represent entities such as video frames, salient objects, and textual cues. These nodes are interconnected via spatial, temporal, and semantic edges, enabling the model to capture fine-grained relationships across modalities. This graph serves as the foundation for context-aware reasoning to improve manipulation action recognition.

Feature representations are extracted using large-scale foundation models: VideoMAE for spatiotemporal video encoding and BERT for contextual text embeddings. The resulting features populate node attributes, while edge attributes encode inter-entity relationships.

Graph reasoning is performed using a Graph Attention Network (GAT), which dynamically modulates message passing based on the relevance of neighboring nodes. This mechanism enables the model to prioritize interactions critical to fine-grained action understanding.

The framework consists of three stages: (1) multimodal feature extraction using VideoMAE and BERT, (2) dynamic graph construction guided by cross-modal cues, and (3) GAT-based reasoning for robust action classification.

We define spatial edges based on patch-level proximity in object embeddings, temporal edges between successive frame nodes, and semantic edges via cosine similarity between text and video features.

\begin{figure}[]
    \centering
    \includegraphics[width=0.95\linewidth]{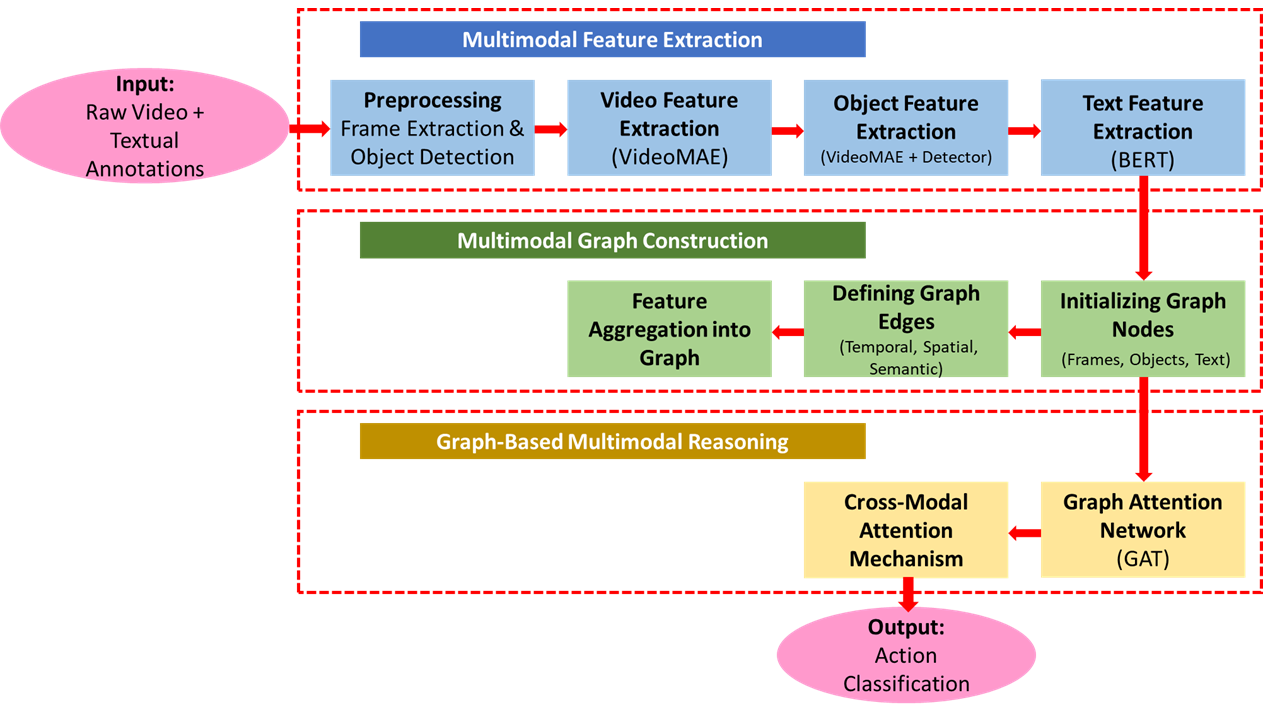} 
    \caption{Overview of the proposed multimodal graph-based framework. 
    The framework consists of three main stages: (1) multimodal feature extraction using VideoMAE for video and BERT for text, (2) multimodal dynamic graph construction where nodes (frames, objects, and text) and edges (temporal, spatial, semantic) are defined, and (3) graph-based multimodal reasoning using Graph Attention Networks (GAT) to refine representations for action classification.}
    \label{fig:framework}
\end{figure}

An overview of the architecture is illustrated in Fig.~\ref{fig:framework}.

\subsection{Multimodal Feature Extraction}

To effectively model bimanual and manipulation actions, we extract \textit{spatio-temporal features} from \textit{video} and \textit{semantic embeddings} from \textit{textual annotations}. This process consists of three key steps: (1) preprocessing, (2) video feature extraction using VideoMAE, and (3) text feature extraction using BERT. Below, we describe these steps.

\subsubsection{\textbf{Preprocessing}}

Before extracting meaningful representations, the raw input data undergoes preprocessing to ensure proper structuring and synchronization of video and text. This step includes:
\begin{itemize}
    \item \textbf{Frame Extraction}: Decomposing video sequences into a series of frames at a fixed frame rate.
    \item \textbf{Text Tokenization}: Processing textual annotations by tokenizing words into meaningful representations.
    \item \textbf{Multimodal Data Synchronization}: Aligning extracted frames with corresponding textual descriptions to ensure consistency across modalities.
\end{itemize}

Let \( V = \{I_1, I_2, \dots, I_T\} \) represent the extracted sequence of frames from a given video, where \( I_t \) is the frame at time \( t \). These frames are stored for subsequent feature extraction without any additional transformations at this stage.

For textual data, we employ the BERT tokenizer, which converts raw text into a token sequence \( T = \{t_1, t_2, \dots, t_n\} \), where each token is mapped to an intermediate representation. The tokenized text is retained and passed directly into the embedding stage.

Since video and text operate at different time scales, we synchronize multimodal data by aligning frames with their corresponding textual annotations. This ensures that extracted features in the next stage maintain proper temporal and semantic correspondence.

The next subsections detail how video and textual embeddings are extracted using Foundation Models and then synchronized.

\subsubsection{\textbf{Video Feature Extraction}}

To obtain robust visual representations for bimanual and manipulation actions, we employ Video Masked Autoencoders (VideoMAE), a transformer-based self-supervised learning model that extracts spatio-temporal embeddings from video sequences. Unlike conventional feature extraction pipelines that rely on hand-crafted descriptors or convolutional architectures, VideoMAE learns a highly generalizable representation through masked frame reconstruction, capturing both local and global motion dynamics. Bimanual actions, characterized by synchronized hand movements and complex object interactions, benefit from VideoMAE's ability to encode both individual and inter-hand dynamics across time.

\paragraph{\textbf{VideoMAE Overview}}
VideoMAE is an extension of Masked Autoencoders (MAE) designed for video data. It applies a high masking ratio to input frames, randomly dropping patches and forcing the model to reconstruct the missing information. This process encourages the network to learn meaningful spatio-temporal patterns, making it highly effective for downstream action recognition tasks. Given an input video sequence \( V = \{I_1, I_2, \dots, I_T\} \), VideoMAE encodes each frame as a set of non-overlapping patches and applies a Vision Transformer (ViT) encoder to extract feature representations.

Formally, let each frame \( I_t \) be divided into \( P \times P \) patches, where each patch is linearly embedded into a feature vector. The masked sequence is then processed by the transformer encoder:
\begin{equation}
    Z_t = \text{ViTEncoder}(\text{Mask}(I_t)),
\end{equation}
where \( Z_t \in \mathbb{R}^{d_V} \) is the encoded feature vector for frame \( I_t \), and \( d_V \) is the feature dimension. The encoder learns to reconstruct missing patches via a lightweight decoder, ensuring that learned embeddings retain both spatial and temporal coherence.\\

\paragraph{\textbf{Extracting Frame-Level Representations}}
For each frame \( I_t \), we extract a deep feature embedding from the final encoder layer of VideoMAE:
\begin{equation}
    \mathbf{f}_v^t = \frac{1}{N} \sum_{i=1}^{N} Z_{t,i},
\end{equation}
where \( \mathbf{f}_v^t \in \mathbb{R}^{d_V} \) represents the aggregated frame embedding, and \( N \) denotes the number of retained visible patches. These frame embeddings serve as \textbf{video frame nodes} in our multimodal graph representation.

\paragraph{\textbf{Extracting Object-Level Representations}}

To capture fine-grained object interactions within video frames, we extract object-centric representations directly from the internal attention maps of VideoMAE. This strategy avoids reliance on explicit object detectors such as YOLO, which require supervised training with extensive bounding box annotations and often generalize poorly to unseen manipulation scenarios. In contrast, VideoMAE’s attention maps implicitly highlight salient regions—such as hands and interacted objects—based on their relevance to masked frame reconstruction. This self-supervised mechanism proves especially advantageous in bimanual tasks, where inter-hand coordination and nuanced object manipulation (e.g., lifting, transferring, assembling) must be captured without manual supervision.

These attention maps assign spatial importance scores to patches within each frame, guiding the model toward regions critical for reconstructing missing content. Notably, these high-attention areas frequently coincide with the objects actively involved in the observed manipulation, enabling robust and adaptive object-level feature extraction.

\paragraph{\textbf{Localized Embedding Extraction}}

For each salient object in a frame, a set of high-attention patches \( \mathcal{R}_o \) is defined. The feature embedding for the object is computed by aggregating the feature vectors of these patches, weighted by their respective attention scores:
\begin{equation}
    \mathbf{f}_o^t = \sum_{i \in \mathcal{R}_o} w_i Z_{t,i}.
\end{equation}
Here:
\begin{itemize}
    \item \( \mathbf{f}_o^t \in \mathbb{R}^{d_V} \) is the feature vector representing the object in frame \( I_t \).
    \item \( \mathcal{R}_o \) is the set of high-attention patch indices associated with the object.
    \item \( Z_{t,i} \in \mathbb{R}^{d_V} \) is the feature vector of the \( i \)-th patch, obtained from VideoMAE.
    \item \( w_i \) is the normalized attention weight for the \( i \)-th patch:
    \[
    w_i = \frac{\text{Attention}(i)}{\sum_{j \in \mathcal{R}_o} \text{Attention}(j)}.
    \]
\end{itemize}

This method produces fine-grained, adaptive object-level embeddings that capture critical details such as the shape, position, and appearance of objects, including hands, tools, and manipulated items. By incorporating these embeddings as object nodes in the multimodal graph, the framework provides a robust basis for reasoning about complex and dynamic actions. The end-to-end nature of this approach ensures adaptability to diverse datasets and tasks, making it highly generalizable.

\paragraph{\textbf{Temporal Feature Aggregation}}
Since manipulation actions unfold over time, we aggregate frame embeddings across multiple time steps to capture long-range dependencies. For bimanual actions, this aggregation is essential to model the temporal synchronization and coordinated interactions between both hands, such as in tasks like pouring water while holding a bottle.

\begin{equation}
    \mathbf{F}_V = \text{TemporalAggregator}(\mathbf{f}_v^1, \mathbf{f}_v^2, \dots, \mathbf{f}_v^T),
\end{equation}
where \( \mathbf{F}_V \in \mathbb{R}^{T \times d_V} \) represents the temporally aggregated video embedding sequence. The aggregator can be instantiated as either a transformer-based self-attention layer or a GAT-style fusion mechanism, allowing the model to capture both local and global temporal dependencies prior to graph construction.

\paragraph{\textbf{Integration with Graph Construction}}
The extracted frame and object embeddings serve as the input nodes of our multimodal graph representation, where each node encapsulates specific aspects of the video content. Frame-level embeddings provide a global scene context for each video frame, capturing spatio-temporal relationships. Object-level embeddings, derived from high-attention regions, offer fine-grained details about salient objects and their interactions within each frame. For bimanual actions, the graph captures relationships between two hands, the objects they manipulate, and their synchronized movements.

\begin{figure}[ht]
    \centering
    \includegraphics[width=\linewidth]{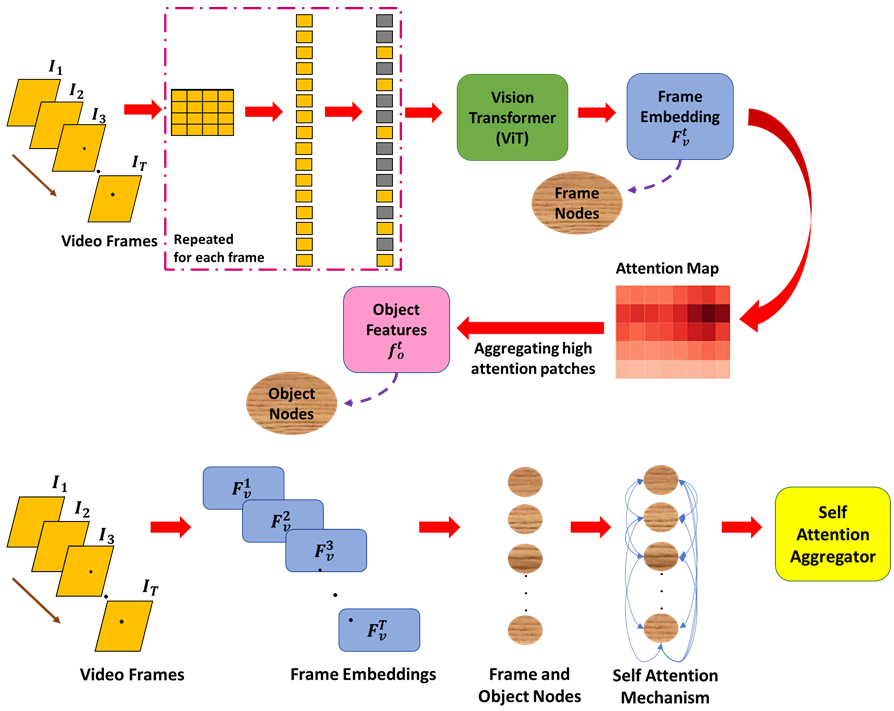} 
    \caption{Overview of the feature extraction and temporal aggregation process. The input video sequence, represented as frames \( \{I_1, I_2, \dots, I_T\} \), is processed by VideoMAE to extract frame embeddings \( \mathbf{f}_v^t \), which are represented as frame nodes. These nodes are connected in a temporal graph, where a self-attention mechanism captures long-range dependencies across time prior to graph-based reasoning. The resulting temporally aggregated embedding summarizes the video sequence for downstream tasks such as action recognition.    }
    \label{video_FE}
\end{figure}

These features are further enhanced by cross-modal attention mechanisms, which dynamically adjust the contributions of different nodes based on their relevance to the task. This process facilitates effective fusion of temporal and spatial information for downstream reasoning. As shown in Figure~\ref{video_FE}, the framework integrates frame nodes and object nodes through a self-attention mechanism, enabling temporal feature aggregation and the modeling of long-range dependencies across the video sequence.
\\

\subsubsection{\textbf{Text Feature Extraction}}

To complement the visual representations extracted from VideoMAE, we process textual annotations to generate semantic embeddings that encode contextual information about actions and objects. These embeddings play a crucial role in enriching the multimodal graph representation by capturing fine-grained relationships between text descriptions, video frames, and object interactions.

\paragraph{\textbf{Input Text Processing.}}
The textual annotations associated with video sequences include labels or descriptive sentences such as ``grasping an object'' or ``pouring water.'' These inputs are tokenized into subword units using a BERT tokenizer, producing a sequence of tokens:
\begin{equation}
    T = \{t_1, t_2, \dots, t_n\},
\end{equation}
where \( t_i \) represents the \( i \)-th token in the sequence. Special tokens such as \texttt{[CLS]} and \texttt{[SEP]} are added to mark the beginning and end of the sequence, respectively, ensuring consistency with the pretrained BERT model architecture.

\paragraph{\textbf{\textbf{Feature Embedding via BERT.}}}
The tokenized sequence \( T \) is passed through a pretrained Bidirectional Encoder Representations from Transformers (BERT) model to compute contextual embeddings for each token:
\begin{equation}
    \mathbf{E} = \{\mathbf{e}_1, \mathbf{e}_2, \dots, \mathbf{e}_n\},
\end{equation}
where \( \mathbf{e}_i \in \mathbb{R}^{d_T} \) is the embedding for the \( i \)-th token, and \( d_T \) denotes the embedding dimensionality. To obtain a single, unified representation for the entire text, we aggregate the token embeddings by selecting the embedding corresponding to the \texttt{[CLS]} token:
\begin{equation}
    \mathbf{f}_t = \mathbf{e}_{\texttt{[CLS]}},
\end{equation}
where \( \mathbf{f}_t \in \mathbb{R}^{d_T} \) serves as the text feature vector representing the input annotation. This vector captures both local and global semantic relationships within the text. BERT generates contextual embeddings that adapt to varied sentence structures and offer strong generalization without task-specific retraining.

\paragraph{\textbf{Integration with Graph Construction}}
The extracted text embeddings \( \mathbf{f}_t \) are incorporated as text nodes within the multimodal graph and connected to frame and object nodes through edges that encode semantic relationships, such as action-object dependencies or temporal alignments. For example, the textual description ``grasping an object'' is linked to relevant object and frame nodes, providing semantic cues to enhance cross-modal reasoning and action differentiation. These interactions are further refined during the graph reasoning stage to ensure accurate action recognition.

\begin{figure}[ht]
    \centering
    \includegraphics[width=0.8\linewidth]{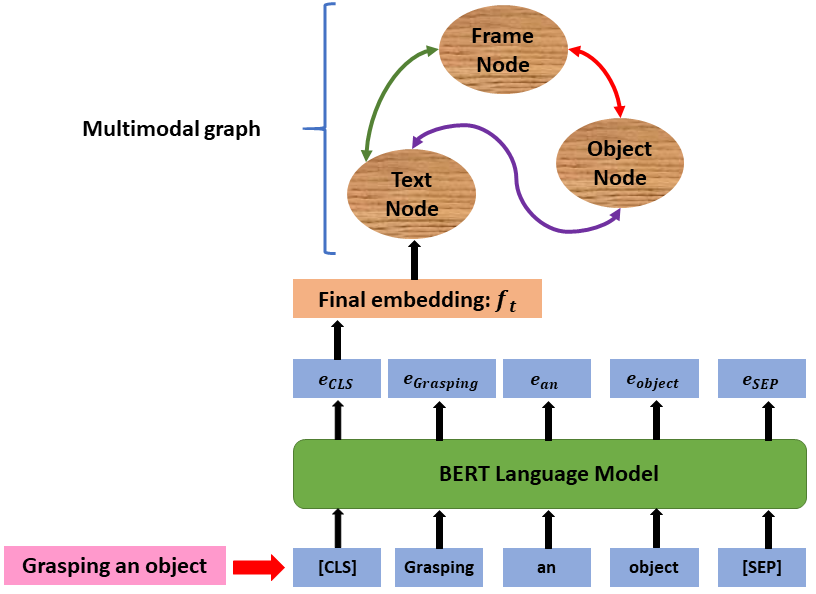} 
    \caption{
        Workflow of text feature extraction using BERT. The input text (e.g., "grasping an object") is tokenized into subwords such as \texttt{[CLS]}, \texttt{grasping}, \texttt{an}, \texttt{object}, and \texttt{[SEP]}. These tokens are passed through the BERT model to generate contextual embeddings for each token. The \texttt{[CLS]} token embedding is aggregated to produce the final text feature vector (\( \mathbf{f}_t \)), which is integrated into the multimodal graph as a text node. This node is connected to frame and object nodes through edges encoding semantic and temporal relationships.
    }
    \label{text_FE}
\end{figure}

As shown in Figure~\ref{text_FE}, the BERT model generates contextual embeddings for text integration.

\subsection{Multimodal Graph Construction}

The core of our proposed framework is the construction of a dynamic multimodal graph \( G = (V, E) \) that unifies visual and textual information within a structured representation. This graph models fine-grained spatial, temporal, and semantic relationships across video frames, salient objects, and corresponding textual descriptions. Here, \( V \) denotes the set of nodes, and \( E \) the set of edges connecting them.

\subsubsection{\textbf{Node Definition}}

The graph contains three types of nodes, each encoding a different modality:

\begin{itemize}
    \item \textbf{Frame Nodes:} Each video frame \( I_t \) is encoded using VideoMAE to produce a feature vector \( \mathbf{f}_v^t \in \mathbb{R}^{d_V} \), representing the global spatiotemporal context of the frame. This embedding serves as the frame node in the graph.

    \item \textbf{Object Nodes:} Salient regions within each frame—such as hands, tools, or manipulated objects—are represented as object nodes. The feature vector for an object node \( \mathbf{f}_o^t \in \mathbb{R}^{d_V} \) is computed by aggregating high-attention patches identified from VideoMAE attention maps:
    \begin{equation}
        \mathbf{f}_o^t = \sum_{i \in \mathcal{R}_o} w_i Z_{t,i},
    \end{equation}
    where \( \mathcal{R}_o \) is the set of high-attention patches associated with the object, \( w_i \) is the normalized attention weight, and \( Z_{t,i} \) is the feature vector for patch \( i \).

    \item \textbf{Text Nodes:} Each textual annotation is represented as a text node, with a feature vector \( \mathbf{f}_t \in \mathbb{R}^{d_T} \) obtained from the embedding of the \texttt{[CLS]} token in a pretrained BERT model:
    \begin{equation}
        \mathbf{f}_t = \mathbf{e}_{\texttt{[CLS]}}.
    \end{equation}
\end{itemize}

\subsubsection{\textbf{Edge Definition}}

Edges encode relationships between nodes and fall into three categories:

\begin{itemize}
    \item \textbf{Temporal Edges:} These connect consecutive frame nodes to capture the progression of motion and interactions across time:
    \begin{equation}
        E_{\text{temporal}} = \{(v_t, v_{t+1}) \mid v_t, v_{t+1} \in V_{\text{frame}}\}.
    \end{equation}

    \item \textbf{Spatial Edges:} These connect object nodes within the same frame, allowing the graph to capture spatial co-occurrence and proximity:
    \begin{equation}
        E_{\text{spatial}} = \{(o_i, o_j) \mid o_i, o_j \in V_{\text{object}},\, i \neq j,\, t_i = t_j\}.
    \end{equation}

    \item \textbf{Semantic Edges:} These connect text nodes to relevant frame or object nodes based on semantic alignment. For instance, a text node describing “grasping an object” connects to hands and the associated frame:
    \begin{equation}
        E_{\text{semantic}} = \{(t, v) \mid t \in V_{\text{text}},\, v \in V_{\text{frame}} \cup V_{\text{object}}\}.
    \end{equation}
\end{itemize}

Although VideoMAE captures high-level spatiotemporal patterns, explicitly modeling temporal edges between frames ensures that fine-grained sequential dependencies are preserved. This is essential for distinguishing manipulation actions that depend on the precise ordering of movements.

\subsubsection{\textbf{Graph Representation}}

The final graph is defined as follows:

\begin{itemize}
    \item \textbf{Node Features:} Each node \( v_i \in V \) is associated with a feature vector:
    \[
    \mathbf{f}_i = 
    \begin{cases} 
      \mathbf{f}_v^t & \text{if } v_i \in V_{\text{frame}}, \\
      \mathbf{f}_o^t & \text{if } v_i \in V_{\text{object}}, \\
      \mathbf{f}_t & \text{if } v_i \in V_{\text{text}}.
    \end{cases}
    \]

    \item \textbf{Edge Weights:} Each edge \( e_{ij} \in E \) is assigned a weight \( w_{ij} \), computed via similarity between node features. For example, cosine similarity can be used:
    \begin{equation}
    w_{ij} = \frac{\mathbf{f}_i \cdot \mathbf{f}_j}{\left\| \mathbf{f}_i \right\| \left\| \mathbf{f}_j \right\|}
\end{equation}
\end{itemize}

\subsubsection{\textbf{Dynamic Graph Construction}}

The multimodal graph is constructed dynamically, adapting to the content of each video sequence and its associated semantic context. This adaptive construction allows the graph to flexibly encode temporal dynamics, spatial relationships, and semantic associations relevant to the manipulation task.

\begin{itemize}
    \item \textbf{Dynamic Node Connectivity:} While temporal edges between frames are fixed to preserve the sequential nature of actions, spatial and semantic edges are instantiated based on pairwise feature similarity. This allows the graph to selectively emphasize relevant object-object and object-text associations in a data-driven manner.
    
    \item \textbf{Edge Weight Refinement:} Edge weights are learned and refined during training, guided by task-specific supervision. The model iteratively adjusts these weights to emphasize interactions most critical for distinguishing between manipulation actions.
\end{itemize}

\begin{figure}[ht]
    \centering
    \includegraphics[width=\linewidth]{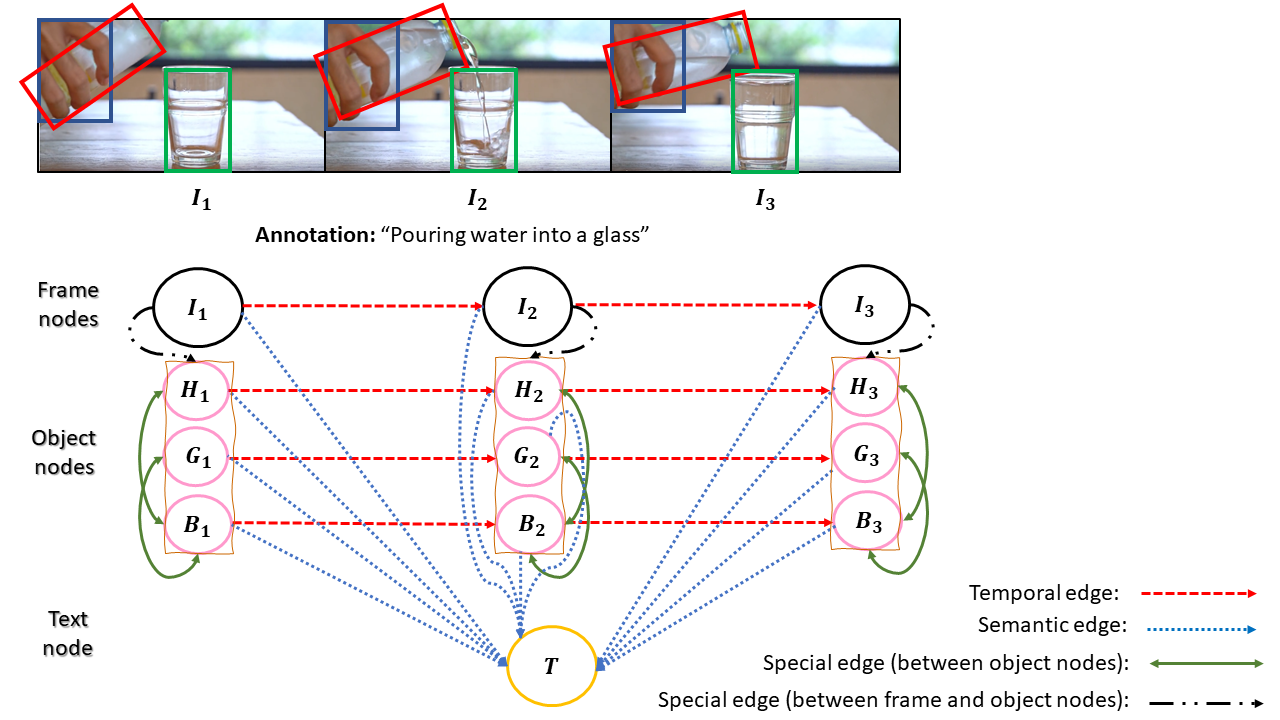} 
    \caption{
        Illustration of the multimodal graph construction process for the action \textit{``Pouring water into a glass.''} 
        The top row shows three frames (\(I_1, I_2, I_3\)) from a video depicting the pouring action. 
        Frame nodes (\(I_1, I_2, I_3\)) represent the temporal context of the video, capturing the sequence of events. 
        Object nodes (\(H\), \(G\), \(B\)) correspond to the hand, glass, and bottle detected in each frame, enabling spatial reasoning. 
        The text node (\(T\)) represents the semantic annotation, providing contextual information about the action. 
        Edges in the graph encode different types of relationships: temporal edges (red dashed lines) connect consecutive frames, spatial edges (green solid and black dotted lines) capture frame-to-object and object-to-object dependencies, and semantic edges (blue dotted lines) link the text node to both frame and object nodes. This multimodal graph structure integrates temporal, spatial, and semantic information for robust action recognition.
    }
    \label{fig:graph_construction}
\end{figure}

Figure~\ref{fig:graph_construction} illustrates this dynamic construction process using the example action \textit{``Pouring water into a glass''}. The multimodal graph consists of frame nodes (\(I_1, I_2, I_3\)), object nodes (e.g., hand (\(H\)), glass (\(G\)), and bottle (\(B\))), and a text node (\(T\)) representing the semantic label.

\paragraph{Temporal and Spatial Connectivity.}
Temporal edges (shown as red dashed arrows) connect consecutive frame nodes in the forward direction, encoding the chronological progression of the video. Spatial edges (solid black arrows) link each frame node to its corresponding object nodes, such as hands and manipulated tools within that frame. These spatial connections flow from frame nodes to object nodes, reflecting the notion that objects are detected and contextualized relative to the frame’s visual content.

\paragraph{Semantic Connectivity.}
Semantic edges (blue dotted arrows) connect object nodes to the text node, capturing how visual entities contribute to the interpretation of the annotated action. For example, the interaction between the hand, glass, and bottle supports the meaning of ``pouring water into a glass.'' These edges are directed from object nodes to the text node, reflecting the data flow from observed visual evidence toward semantic interpretation. By contrast, edges are not drawn from the text node to visual elements, as the textual description does not influence object behavior but rather summarizes observed patterns.

This directional edge design ensures that the graph architecture mirrors the natural information flow in video-based action understanding—from visual observation to semantic abstraction.

\subsubsection{\textbf{Synchronization for Multimodal Alignment}}

To ensure coherent multimodal representation, synchronization aligns temporal and semantic information across visual and textual streams. Each video frame is temporally mapped to its corresponding textual annotation (e.g., \textit{``pouring water into a glass''}), while object-level features are anchored to their source frames to preserve spatial and temporal locality.

This alignment enables the graph to maintain a consistent frame order—crucial for modeling sequential dependencies via temporal edges—and to accurately associate semantic cues with relevant visual entities. As a result, the constructed multimodal graph supports joint reasoning over time, space, and meaning.

\subsection{Graph-Based Multimodal Reasoning}
Graph-based reasoning is a fundamental step in our framework, enabling the integration of temporal, spatial, and semantic information to recognize complex bimanual and manipulation actions. By leveraging a graph-based approach, we can represent interactions between video frames, objects, and textual annotations in a unified structure. The Graph Attention Network (GAT) serves as the backbone for our reasoning, allowing the model to focus on task-specific relationships between nodes and adapt dynamically to the evolving nature of these interactions. This reasoning step ensures the effective aggregation and refinement of multimodal features, facilitating precise action recognition.

\subsubsection{\textbf{Task-Specific Attention Mechanisms}}

Recognizing bimanual and manipulation actions requires a model that can dynamically prioritize relationships among video frames, object regions, and textual annotations. To this end, we introduce a task-specific attention mechanism within the Graph Attention Network (GAT), which adaptively focuses on the most contextually relevant interactions based on both node features and domain-specific cues.

Let \( \mathbf{h}_i \) and \( \mathbf{h}_j \) denote the current-layer embeddings of nodes \( v_i \) and \( v_j \), respectively, as produced by the previous GAT layer. The attention weight \( \alpha_{ij} \) for edge \( e_{ij} \) is computed as:
\begin{equation}
    \alpha_{ij} = \frac{\exp\left(\text{LeakyReLU}\left(\mathbf{a}^\top \left[\mathbf{W}\mathbf{h}_i \| \mathbf{W}\mathbf{h}_j \right]\right) + \phi(i, j)\right)}{\sum_{k \in \mathcal{N}(i)} \exp\left(\text{LeakyReLU}\left(\mathbf{a}^\top \left[\mathbf{W}\mathbf{h}_i \| \mathbf{W}\mathbf{h}_k \right]\right) + \phi(i, k)\right)},
\end{equation}
where \( \mathbf{W} \) is a learnable weight matrix, \( \mathbf{a} \) is a shared attention vector, \( \| \) denotes concatenation, and \( \phi(i, j) \) is a task-specific modulation term that embeds prior knowledge into the attention computation.

The modulation term \( \phi(i, j) \) adjusts edge importance based on the edge type and its relevance to the manipulation task:
\begin{equation}
    \phi(i, j) = 
    \begin{cases} 
      \omega_{\text{temporal}} \cdot \delta(i, j), & \text{if } e_{ij} \text{ is a temporal edge}, \\
      \omega_{\text{spatial}} \cdot \text{Sim}(\mathbf{h}_i, \mathbf{h}_j), & \text{if } e_{ij} \text{ is a spatial edge}, \\
      \omega_{\text{semantic}} \cdot f_{\text{sem}}(i, j), & \text{if } e_{ij} \text{ is a semantic edge}.
    \end{cases}
\end{equation}
Here, \( \omega_{\text{temporal}}, \omega_{\text{spatial}}, \omega_{\text{semantic}} \) are learnable scalars that control the contribution of each edge type. The function \( \delta(i, j) \) ensures temporal consistency by emphasizing sequential adjacency, while \( f_{\text{sem}}(i, j) \) is a trainable compatibility function that models semantic relevance between textual and visual nodes.

We define \( \text{Sim}(\mathbf{h}_i, \mathbf{h}_j) \) as the cosine similarity between spatially connected node embeddings:
\[
\text{Sim}(\mathbf{h}_i, \mathbf{h}_j) = \frac{\mathbf{h}_i \cdot \mathbf{h}_j}{\|\mathbf{h}_i\| \|\mathbf{h}_j\|}.
\]

The learned attention weights \( \alpha_{ij} \) guide message passing within the graph. Node representations are updated layer-wise as:
\begin{equation}
    \mathbf{h}_i^{(l+1)} = \sigma\left(\sum_{j \in \mathcal{N}(i)} \alpha_{ij}^{(l)} \mathbf{W}^{(l)} \mathbf{h}_j^{(l)}\right),
\end{equation}
where \( \mathbf{h}_i^{(l)} \) denotes the feature of node \( v_i \) at layer \( l \), and \( \sigma(\cdot) \) is a nonlinear activation function such as ReLU.

Embedding task-specific priors into the attention mechanism offers several benefits: (i) it enables context-aware reasoning by emphasizing relationships critical to recognizing subtle inter-hand and hand-object interactions; (ii) it adapts edge importance based on both input content and task semantics; and (iii) it promotes generalization to unseen action types through principled integration of multimodal and structural information.

\subsubsection{\textbf{Dynamic Graph Adaptation}}

Extending the dynamic construction process introduced earlier, this component focuses on how the graph structure is iteratively refined during training. While the initial connectivity is established using cross-modal cues, our model continuously updates edge weights and topologies based on task-specific objectives. This facilitates precise modeling of spatiotemporal and semantic dependencies critical to bimanual manipulation actions.

Importantly, while the modulation term \( \phi(i, j) \) introduced earlier influences how node features are aggregated during attention, the adjustment term \( \psi(i, j) \) directly modifies edge weights during training. This distinction enables our framework to adapt both the message-passing dynamics and the underlying graph topology in a task-aware manner.

\paragraph{\textbf{Dynamic Edge Weighting}}
To adaptively model the importance of relationships, we introduce a dynamic edge weighting mechanism. Each edge \( e_{ij} \in E \) is assigned a weight \( w_{ij} \), which is iteratively updated based on the similarity between node features and task-specific relevance. The edge weight is computed as:
\begin{equation}
    w_{ij} = \text{Sim}(\mathbf{h}_i, \mathbf{h}_j) + \psi(i, j),
\end{equation}
where \( \text{Sim}(\mathbf{h}_i, \mathbf{h}_j) \) measures the similarity between the feature embeddings of nodes \( v_i \) and \( v_j \), and \( \psi(i, j) \) is a task-specific adjustment term that accounts for the type and context of the edge. This adjustment term incorporates domain-specific knowledge, allowing the model to prioritize edges that are most relevant for recognizing complex actions.

\paragraph{\textbf{Edge Refinement During Training}}
During training, the edge weights are iteratively refined through backpropagation to align with the task objectives. For temporal edges, the model emphasizes consistency between consecutive frames, while spatial edges are refined to highlight critical hand-object interactions. Semantic edges are dynamically adjusted to strengthen the connections between textual descriptions and their corresponding visual elements.

Formally, the task-specific adjustment term \( \psi(i, j) \) is defined as:
\begin{equation}
    \psi(i, j) = 
    \begin{cases} 
      \lambda_{\text{temporal}} \cdot \delta(i, j), & \text{if } e_{ij} \text{ is a temporal edge}, \\
      \lambda_{\text{spatial}} \cdot \text{Sim}(\mathbf{h}_i, \mathbf{h}_j), & \text{if } e_{ij} \text{ is a spatial edge}, \\
      \lambda_{\text{semantic}} \cdot f_{\text{sem}}(i, j), & \text{if } e_{ij} \text{ is a semantic edge}.
    \end{cases}
\end{equation}
Here, \( \lambda_{\text{temporal}}, \lambda_{\text{spatial}}, \lambda_{\text{semantic}} \) are learnable parameters that balance the contribution of different edge types. The functions \( \delta(i, j) \), \( \text{Sim}(\mathbf{h}_i, \mathbf{h}_j) \), and \( f_{\text{sem}}(i, j) \) capture temporal consistency, spatial similarity, and semantic relevance, respectively.

\paragraph{\textbf{Dynamic Connectivity}}
In addition to updating edge weights, our framework supports dynamic connectivity by pruning or adding edges during training. Edges with weights below a threshold are pruned to reduce noise, while new edges are formed between nodes with high feature similarity or strong semantic alignment. This dynamic connectivity ensures that the graph structure remains flexible and adapts to the underlying data distribution.

\paragraph{\textbf{Adaptive Graph Representation}}
At each training iteration, the graph representation is updated based on the refined edge weights and connectivity. The updated graph serves as input to the Graph Attention Network (GAT), which propagates information across nodes using the dynamically adapted structure. This process enhances the model’s ability to capture subtle interactions, such as synchronized hand movements or object manipulations, that are critical for bimanual action recognition.

\subsubsection{\textbf{Multimodal Feature Fusion}}

Multimodal feature fusion integrates spatiotemporal video embeddings with semantic textual cues to create a unified representation for reasoning about bimanual and manipulation actions. This fusion is enhanced through task-specific attention mechanisms and dynamic graph adaptation, ensuring context-aware and structurally flexible modeling.

\paragraph{\textbf{Fusion via Task-Specific Attention}}
To prioritize modality-specific relevance, we project the video and text embeddings into a shared space:
\begin{equation}
    \mathbf{f}_v' = \mathbf{W}_v \mathbf{f}_v, \quad \mathbf{f}_t' = \mathbf{W}_t \mathbf{f}_t,
\end{equation}
where \( \mathbf{W}_v \) and \( \mathbf{W}_t \) are learnable linear projections. Task-specific attention weights are computed as:
\begin{equation}
    \alpha_v = \text{softmax}(\mathbf{a}^\top \mathbf{f}_v'), \quad \alpha_t = \text{softmax}(\mathbf{a}^\top \mathbf{f}_t'),
\end{equation}
using a shared attention vector \( \mathbf{a} \). The final fused feature is:
\begin{equation}
    \mathbf{f}_{\text{fusion}} = \alpha_v \mathbf{f}_v' + \alpha_t \mathbf{f}_t'.
\end{equation}

\paragraph{\textbf{Graph Integration and Adaptation}}
The fused feature \( \mathbf{f}_{\text{fusion}} \) is embedded as a node in the multimodal graph. Task-specific attention and dynamic graph adaptation guide edge formation and weighting: temporal edges track action progression, spatial edges model inter-object and hand-object coordination, and semantic edges align language cues with visual evidence. Through iterative refinement during training, the graph dynamically evolves to emphasize the most informative relationships.

\begin{figure}[ht]
    \centering
    \includegraphics[width=0.95\linewidth]{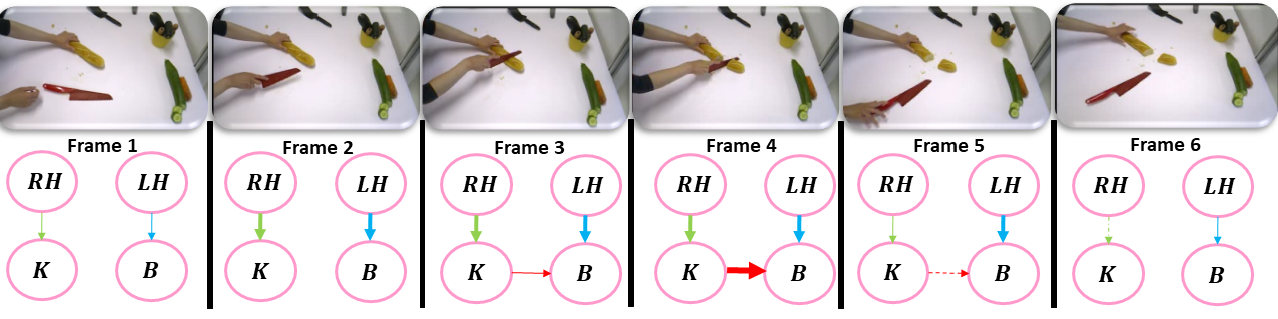}
    \caption{\textbf{Task-Specific Attention and Dynamic Graph Adaptation in Bimanual Manipulation (Cutting Bread).}  
This figure shows evolving hand-object and object-object interactions across six frames. Nodes represent the right hand (RH), left hand (LH), knife (K), and bread (B). Edges adapt in strength based on contextual relevance as the cutting action progresses.}
    \label{reasoning}
\end{figure}

Figure~\ref{reasoning} illustrates this adaptive reasoning process using a six-frame cutting sequence. As the action unfolds, edge strengths evolve to highlight key interactions (e.g., knife–bread), reflecting the model’s ability to capture fine-grained manipulation dynamics.

After multimodal reasoning, final node embeddings are aggregated using global mean pooling to form a graph-level representation. This is passed through a fully connected layer for action classification. The full model—spanning feature extraction, graph construction, and reasoning—is trained end-to-end via cross-entropy loss, enabling optimized recognition of complex bimanual manipulation behaviors.

\section{Results}
\label{sec:results}

\subsection{Experimental Setup}

\subsubsection{\textbf{Datasets}}
We evaluate our framework on four publicly available benchmarks that feature both video content and aligned textual annotations, enabling comprehensive multimodal evaluation:

\begin{itemize}
    \item \textbf{EPIC-KITCHENS}~\cite{Damen2018EPICKITCHENS}: A 55-hour egocentric video dataset of daily kitchen activities with dense action annotations and narrations, enabling fine-grained hand-object interaction analysis.
    \item \textbf{YouCookII}~\cite{zhou2018towards}: A collection of 2,000 cooking videos annotated with temporal boundaries and free-form descriptions, supporting video-text alignment.
    \item \textbf{Something-Something V2}~\cite{goyal2017something}: Contains 220,000 short clips emphasizing object manipulation, annotated with templated textual phrases, facilitating fine-grained spatio-temporal reasoning.
    \item \textbf{COIN}~\cite{tang2019coin}: Includes 11,000 instructional videos across 180 tasks, with step-by-step segmentation and text, supporting procedural action understanding.
\end{itemize}

\subsubsection{\textbf{Implementation Details}}
We implement our framework in PyTorch and train it on a single NVIDIA A100 GPU (40 GB). Visual features are extracted using a pretrained VideoMAE (768-dimensional), while semantic embeddings are obtained from BERT-base (also 768-dimensional). These embeddings are projected into a shared feature space and used to initialize the multimodal graph. The entire architecture, including graph-based reasoning, is trained end-to-end for action classification.

\subsubsection{\textbf{Training Procedure}}
We optimize the model using the Adam optimizer with an initial learning rate of \(1 \times 10^{-4}\), following a cosine annealing schedule with a five-epoch warm-up. Training is conducted for 50 epochs with a batch size of 32. Cross-entropy loss is used to supervise action classification, and all hyperparameters are selected based on validation set performance.

\subsection{Comparison with State-of-the-Art Methods}
In this section, we evaluate our proposed multimodal graph-based framework by comparing it with state-of-the-art approaches. Specifically, we conduct detailed comparisons focusing on four key aspects: (1) \textit{Recognition Accuracy}, assessing improvements in top-1 and top-5 accuracy; (2) \textit{Generalization Capability}, examining the performance on novel or unseen manipulation actions; (3) \textit{Multimodal Fusion Effectiveness}, analyzing the benefits derived from integrating visual and textual modalities; and (4) \textit{Graph Structure Flexibility and Adaptability}, highlighting the advantages of our dynamic graph adaptation and task-specific attention mechanisms compared to existing static and adaptive graph methods. Additionally, we present an ablation study to quantify the individual contributions of core components within our framework.

\subsubsection{\textbf{Recognition Accuracy}}

Recognition accuracy serves as a primary benchmark for evaluating action recognition systems, especially in tasks involving fine-grained manipulation dynamics. Table~\ref{tab:recognition_metrics} presents a comparative analysis of our proposed method against recent state-of-the-art approaches across four widely-used datasets.

Our model consistently outperforms existing methods by jointly leveraging foundation model-based multimodal fusion and dynamic graph reasoning. Specifically, the integration of VideoMAE (for spatio-temporal visual encoding) and BERT (for semantic text representation) enables robust disambiguation of visually similar actions. Furthermore, our adaptive graph refinement mechanism enhances the modeling of object-centric and inter-hand interactions, leading to improved classification accuracy across diverse manipulation scenarios.

\begin{table*}[htbp]
\centering
\caption{Recognition Accuracy, Precision, Recall, and F1 Score (\%) Comparison Across Datasets}
\label{tab:recognition_metrics}
\resizebox{\textwidth}{!}{%
\begin{tabular}{lcccccccccccccccc}
\toprule
\textbf{Method} & \multicolumn{4}{c}{\textbf{Accuracy}} & \multicolumn{4}{c}{\textbf{Precision}} & \multicolumn{4}{c}{\textbf{Recall}} & \multicolumn{4}{c}{\textbf{F1 Score}} \\
\cmidrule(lr){2-5} \cmidrule(lr){6-9} \cmidrule(lr){10-13} \cmidrule(lr){14-17}
& EK & SSv2 & YCII & COIN & EK & SSv2 & YCII & COIN & EK & SSv2 & YCII & COIN & EK & SSv2 & YCII & COIN \\
\midrule
H2OTR~\cite{cho2023transformer} & 65.2 & 69.7 & -- & -- & 66.1 & 70.4 & -- & -- & 64.0 & 69.0 & -- & -- & 65.0 & 69.7 & -- & -- \\
HandFormer~\cite{shamil2024utility} & 67.1 & 72.3 & -- & -- & 68.0 & 73.1 & -- & -- & 66.2 & 71.5 & -- & -- & 67.1 & 72.3 & -- & -- \\
Fusion-GCN~\cite{duhme2021fusiongcn} & 60.8 & 67.9 & 62.4 & 72.6 & 61.2 & 68.5 & 63.0 & 71.8 & 60.0 & 67.1 & 61.8 & 73.2 & 60.6 & 67.8 & 62.4 & 72.5 \\
ActionCLIP~\cite{wang2023actionclip} & 66.3 & 71.5 & 64.8 & 74.1 & 66.8 & 72.1 & 65.2 & 73.5 & 65.5 & 70.9 & 64.5 & 74.8 & 66.1 & 71.5 & 64.8 & 74.1 \\
RDT~\cite{liu2025rdt} & 66.7 & 72.1 & 64.3 & 74.5 & 67.3 & 72.8 & 64.9 & 74.0 & 66.0 & 71.3 & 63.7 & 75.2 & 66.6 & 72.0 & 64.3 & 74.6 \\
\textbf{Ours} & \textbf{69.8} & \textbf{75.4} & \textbf{68.1} & \textbf{77.2} & \textbf{70.5} & \textbf{76.2} & \textbf{68.7} & \textbf{76.8} & \textbf{69.0} & \textbf{74.5} & \textbf{67.4} & \textbf{77.8} & \textbf{69.7} & \textbf{75.3} & \textbf{68.0} & \textbf{77.3} \\
\bottomrule
\end{tabular}%
}
\end{table*}

\textit{Note: H2OTR and HandFormer were not evaluated on YouCookII and COIN in their original works.}

\subsubsection{\textbf{Generalization Capability}}
Evaluating the generalization capability of action recognition models is crucial to understanding their robustness in real-world scenarios. We assess generalization using three key evaluations: 
(1) \textit{Cross-dataset transfer}, where models are trained on one dataset and tested on another; 
(2) \textit{Few-shot learning}, which measures performance when limited labeled samples are available; and 
(3) \textit{Unseen class recognition}, where certain action categories are excluded during training and evaluated separately.

Tables ~\ref{tab:cross_dataset_transfer} to ~\ref{tab:unseen_class_recognition} presents a comparison of our approach with recent state-of-the-art methods on generalization tasks.

\begin{table*}[htbp]
\centering
\caption{Cross-Dataset Transfer Accuracy (\%) Comparison}
\label{tab:cross_dataset_transfer}
\renewcommand{\arraystretch}{1.3} 
\resizebox{\textwidth}{!}{%
\begin{tabular}{lcccc}
\hline
\textbf{Method} & \shortstack{\textbf{EPIC-KITCHENS} \\ $\rightarrow$ \\ \textbf{Something-Something V2}} & 
\shortstack{\textbf{Something-Something} \\ $\rightarrow$ \\ \textbf{YouCookII}} & 
\shortstack{\textbf{YouCookII} \\ $\rightarrow$ \\ \textbf{COIN}} & 
\textbf{Average} \\
\hline
HAT~\cite{liu2023hat} & 52.4 & 49.8 & 54.1 & 52.1 \\
CDAN~\cite{li2023cross} & 53.7 & 50.9 & 55.3 & 53.3 \\
AST-GCN~\cite{chen2023adaptive} & 55.1 & 52.6 & \textbf{58.3} & 55.3 \\
TARN~\cite{li2024transferring} & 54.3 & 53.1 & 57.2 & 54.9 \\
\textbf{Proposed Method} & \textbf{56.4} & \textbf{54.0} & 57.9 & \textbf{56.1} \\
\hline
\end{tabular}%
}
\end{table*}

\begin{table}[htbp]
\centering
\caption{Few-Shot Recognition Accuracy (\%) on Manipulation Actions}
\label{tab:few_shot_recognition}
\renewcommand{\arraystretch}{1.2} 
\begin{tabular}{lcccc}
\hline
\textbf{Method} & \textbf{1-Shot} & \textbf{5-Shot} & \textbf{10-Shot} & \textbf{Average} \\ \hline
ProtoNet~\cite{snell2017prototypical} & 38.2 & 49.5 & 56.1 & 47.9 \\
MatchingNet~\cite{vinyals2016matching} & 40.4 & 51.3 & 57.8 & 49.8 \\
RelationNet~\cite{sung2018learning} & 42.9 & 53.6 & 59.1 & 51.9 \\
TARN~\cite{li2024transferring} & \textbf{45.9} & 56.7 & 62.3 & 54.9 \\
\textbf{Proposed Method} & 43.8 & \textbf{58.9} & \textbf{65.2} & \textbf{56.0} \\ \hline
\end{tabular}
\end{table}

\begin{table}[htbp]
\centering
\caption{Unseen Class Recognition Accuracy (\%) in Zero-Shot Settings}
\label{tab:unseen_class_recognition}
\begin{tabular}{lccc}
\hline
\textbf{Method} & \textbf{Seen Classes} & \textbf{Unseen Classes} & \textbf{Harmonic Mean (H-Mean)} \\ \hline
DeViSE~\cite{frome2013devise} & 72.5 & 34.2 & 46.5 \\
ActionCLIP~\cite{wang2023actionclip} & 74.3 & 38.1 & \textbf{50.4} \\
ZS-GCN~\cite{gao2023zero} & \textbf{77.1} & 40.7 & 52.8 \\
\textbf{Proposed Method} & 76.8 & \textbf{42.5} & 52.3 \\ \hline
\end{tabular}
\end{table}

Our model demonstrates strong generalization capabilities across diverse evaluation settings, outperforming existing methods in most cases while exhibiting room for improvement in certain scenarios. 

In \textit{cross-dataset transfer} (Table~\ref{tab:cross_dataset_transfer}), our approach achieves the highest average accuracy by effectively leveraging foundation models and adaptive graph reasoning to transfer knowledge between datasets with varying distributions. However, AST-GCN~\cite{chen2023adaptive} slightly surpasses our model in the YouCookII $\rightarrow$ COIN transfer task, indicating its superior ability to capture instructional video dependencies.

For \textit{few-shot recognition} (Table~\ref{tab:few_shot_recognition}), our method excels in 5-shot and 10-shot settings, benefiting from task-specific attention and multimodal fusion, which enhance representation learning with limited supervision. However, TARN~\cite{li2024transferring} achieves slightly better performance in the 1-shot setting, suggesting its stronger adaptability in extreme low-data regimes.

In \textit{unseen class recognition} (Table~\ref{tab:unseen_class_recognition}), our method achieves the highest accuracy on unseen classes, demonstrating its robustness in generalizing to novel actions. However, ZS-GCN~\cite{gao2023zero} achieves a slightly higher harmonic mean, likely due to its explicit graph-based semantic alignment for zero-shot tasks.

Overall, our approach effectively balances transferability, adaptability, and robustness by integrating multimodal foundation models with dynamic graph adaptation. While some state-of-the-art methods excel in specific cases, our model provides a well-rounded solution, achieving consistent performance across generalization benchmarks.

\subsubsection{\textbf{Multimodal Fusion Effectiveness}}

The effectiveness of multimodal fusion is a key factor in improving action recognition, as integrating video and textual information enhances contextual understanding. However, aligning heterogeneous modalities remains challenging due to varying temporal structures and semantic gaps between vision and language. This section evaluates our method's fusion effectiveness compared to existing approaches.

\begin{table}[htbp]
\centering
\caption{Improvements from Multimodal Fusion over a Visual-Only Baseline (VideoMAE) on EPIC-KITCHENS. Gains are shown in percentage points across evaluation metrics.}
\label{tab:multimodal_fusion_improvement}
\renewcommand{\arraystretch}{1.2}
\resizebox{\linewidth}{!}{%
\begin{tabular}{lcccc l}
\hline
\textbf{Method} & \textbf{Accuracy Gain (\%)} & \textbf{Precision Gain (\%)} & \textbf{Recall Gain (\%)} & \textbf{F1 Gain (\%)} & \textbf{Fusion Strategy} \\
\hline
MUREL~\cite{cadene2019murel} & 3.1 & 2.8 & 2.5 & 2.6 & Attention-based \\
ClipBERT~\cite{lei2021less} & 4.5 & 3.7 & 3.9 & 3.4 & Late Fusion \\
UniVL~\cite{luo2021univl} & 5.2 & 4.2 & 4.0 & 4.1 & Early Fusion \\
ActionCLIP~\cite{wang2023actionclip} & 7.0 & 6.3 & 5.7 & 5.9 & Text-Visual Alignment \\
\textbf{Proposed Method} & \textbf{7.7} & \textbf{6.9} & \textbf{6.5} & \textbf{6.7} & Adaptive Attention Fusion \\
\hline
\end{tabular}
}
\end{table}

Our method outperforms existing fusion strategies by leveraging adaptive attention fusion, dynamically refining the contribution of each modality based on task relevance. Unlike early or late fusion approaches, our method ensures context-aware alignment, mitigating discrepancies between textual descriptions and video features. While ActionCLIP achieves strong text-visual alignment, it lacks fine-grained adaptation, which limits its effectiveness in complex manipulation actions. MUREL~\cite{cadene2019murel} and UniVL~\cite{luo2021univl} focus on multimodal relational reasoning but primarily optimize for general vision-language tasks, making them less effective in modeling dynamic hand-object interactions. Similarly, ClipBERT~\cite{lei2021less} adopts sparse video sampling, which reduces computational overhead but overlooks fine-grained motion details essential for precise action segmentation. In contrast, our method integrates VideoMAE for dense spatio-temporal feature extraction and BERT for textual semantics, while dynamically refining multimodal relationships using graph-based reasoning. This enables more robust fusion, ensuring precise action recognition even in ambiguous or visually similar tasks.

\subsubsection{\textbf{Graph Structure Flexibility and Adaptability}}

A key advantage of our approach lies in its capacity to dynamically adapt graph structures based on contextual information. Unlike models that rely on static graph topologies, our framework adjusts node connectivity during inference, enabling it to more accurately capture evolving hand-object interactions.

Traditional methods like ST-GCN~\cite{yan2018spatial} and 2s-AGCN~\cite{shi2019two} employ predefined graph structures, limiting their adaptability to dynamic scenarios. While AST-GCN~\cite{chen2023adaptive} introduces adaptive graph learning, its focus on skeletal data restricts its effectiveness in modeling fine-grained object interactions inherent in manipulation tasks.

\begin{table}[htbp]
\centering
\caption{Comparison of Graph Structure Adaptability Across Methods}
\label{tab:graph_flexibility}
\renewcommand{\arraystretch}{1.2}
\begin{tabular}{lccc}
\hline
\textbf{Method} & \textbf{Static Graph} & \textbf{Adaptive Edge Weights} & \textbf{Dynamic Topology} \\
\hline
ST-GCN~\cite{yan2018spatial} & \checkmark & \xmark & \xmark \\
2s-AGCN~\cite{shi2019two} & \checkmark & \checkmark & \xmark \\
AST-GCN~\cite{chen2023adaptive} & \xmark & \checkmark & Limited \\
BiGNN~\cite{Ziaeetabar2024} & \checkmark & \checkmark & \xmark \\
\textbf{Proposed Method} & \xmark & \checkmark & \checkmark \\
\hline
\end{tabular}
\end{table}

Our method incorporates task-specific attention mechanisms that refine node connectivity based on action semantics, emphasizing critical interactions while pruning redundant connections. This adaptability allows the model to generalize effectively across diverse manipulation contexts, leading to improved performance in dynamic environments.

\subsubsection{\textbf{Ablation Study}}

To assess the contribution of each core component in our proposed framework, we conduct an ablation study on the EPIC-KITCHENS dataset. This analysis quantifies the individual and combined effects of multimodal integration, graph-based reasoning, and dynamic topology adaptation. We report Top-1 accuracy and F1 score as evaluation metrics.

\begin{table}[htbp]
\centering
\caption{Ablation Study on EPIC-KITCHENS. Each component is incrementally added to assess its impact.}
\label{tab:ablation}
\renewcommand{\arraystretch}{1.2}
\resizebox{\linewidth}{!}{%
\begin{tabular}{lccccc}
\hline
\textbf{Variant} & \textbf{VideoMAE} & \textbf{BERT} & \textbf{Graph Reasoning} & \textbf{Dynamic Topology} & \textbf{Accuracy / F1 (\%)} \\
\hline
Visual Baseline (VideoMAE only) & \checkmark & \xmark & \xmark & \xmark & 59.1 / 58.2 \\
+ Text (Late Fusion) & \checkmark & \checkmark & \xmark & \xmark & 61.7 / 60.8 \\
+ Static Graph Reasoning & \checkmark & \checkmark & \checkmark & \xmark & 63.1 / 62.0 \\
\textbf{Full Model (Ours)} & \checkmark & \checkmark & \checkmark & \checkmark & \textbf{69.8 / 68.0} \\
\hline
\end{tabular}
}
\end{table}
The results in Table~\ref{tab:ablation} highlight several key insights:
\begin{itemize}
    \item Incorporating BERT-based textual embeddings improves recognition by 2.6\% in accuracy, demonstrating the benefit of semantic context in action disambiguation.
    \item Adding graph-based reasoning (with static topology) yields further gains, underscoring the value of modeling inter-entity relationships.
    \item Enabling dynamic topology refinement leads to the best performance, confirming that adaptive structural updates allow the model to better capture evolving hand-object interactions.
\end{itemize}

\section{Conclusion and Future Work}

In this paper, we proposed a novel multimodal graph-based framework for fine-grained manipulation action recognition, integrating the representational power of foundation models with adaptive graph reasoning. By coupling VideoMAE for spatiotemporal video encoding and BERT for contextual textual understanding, our model constructs a dynamic multimodal graph where node connectivity and edge importance evolve based on task-specific cues. This allows for effective modeling of complex hand-object and inter-hand interactions across time and semantics.

Comprehensive evaluations across four benchmark datasets—EPIC-KITCHENS, Something-Something V2, YouCookII, and COIN—demonstrate that our framework consistently outperforms existing methods in recognition accuracy, generalization capability, and fusion effectiveness. Additionally, our dynamic graph structure introduces greater flexibility and adaptability compared to traditional static or partially adaptive models.

Despite these promising results, some limitations remain. The reliance on large-scale foundation models incurs high computational overhead, which may hinder real-time or embedded deployments. Moreover, while our model demonstrates strong cross-dataset transferability, scalability to extremely long action sequences or highly diverse multimodal data still warrants further exploration.

\textbf{Future work} will focus on three key directions: (1) developing lightweight or distilled variants of the model for real-time applications; (2) extending the framework to incorporate additional modalities such as audio and depth for richer contextual understanding; and (3) investigating self-supervised and few-shot learning strategies to enhance performance in low-resource and cross-domain settings. These improvements aim to broaden the applicability and efficiency of multimodal graph-based action recognition in real-world environments.

\section{Acknowledgment}
This publication was funded by the Deutsche Forschungsgemeinschaft (DFG, German Research Foundation) - Project-ID 454648639 - SFB 1528 (Project Z02)

\bibliographystyle{ACM-Reference-Format}
\bibliography{References}

%
%
%

\end{document}